\documentclass[12pt]{article}

\usepackage[margin=1.25in]{geometry}  
\usepackage{titlesec}  
\usepackage{setspace}  
\usepackage{graphicx}  
\usepackage{caption}   
\usepackage{fancyhdr}  
\usepackage{graphicx}

\usepackage{amsmath,amssymb,amsfonts,mathtools,amsthm}
\usepackage{graphicx}
\usepackage{textcomp}
\usepackage{xcolor}
\usepackage{booktabs}
\usepackage{algorithm}
\usepackage{algpseudocode}

\usepackage[english]{babel}
\usepackage{natbib}
\titleformat{\title}{\centering\Huge\bfseries}{}{0pt}{}
\titleformat{\author}{\centering\large}{}{0pt}{}

\titleformat{\section}{\large\bfseries}{\thesection}{1em}{}
\titleformat{\subsection}{\normalsize\bfseries}{\thesubsection}{1em}{}
\titleformat{\subsubsection}{\normalsize\bfseries}{\thesubsubsection}{1em}{}

\title{AI Driven Soccer Analysis Using Computer Vision}

\author{
    Adrian Manchado, Tanner Cellio,  Jonathan Keane, \\ Yiyang Wang \\
       Department of Computer Science and Software Engineering\\
       Milwaukee School of Engineering\\
       Milwaukee, WI 53202
       \\
       \{manchadoa, celliot, keanej, wangyi\}@msoe.edu}

\date{March 22, 2025}

\begin{document}


\maketitle

\thispagestyle{empty}

\begin{abstract}
Sport analysis is crucial for team performance since it provides actionable data that can inform coaching decisions, improve player performance, and enhance team strategies. To analyze more complex features from game footage, a computer vision model can be used to identify and track key entities from the field. We propose the use of an object detection and tracking system to predict player positioning throughout the game. To translate this to positioning in relation to the field dimensions, we use a point prediction model to identify key points on the field and combine these with known field dimensions to extract actual distances. For the player-identification model, object detection models like YOLO and Faster R-CNN are evaluated on the accuracy of our custom video footage using multiple different evaluation metrics. The goal is to identify the best model for object identification to obtain the most accurate results when paired with SAM2 (Segment Anything Model 2) for segmentation and tracking. For the key point detection model, we use a CNN model to find consistent locations in the soccer field.  Through homography, the positions of points and objects in the camera perspective will be transformed to a real-ground perspective. The segmented player masks from SAM2 are transformed from camera perspective to real-world field coordinates through homography, regardless of camera angle or movement. The transformed real-world coordinates can be used to calculate valuable tactical insights including player speed, distance covered, positioning heatmaps, and more complex team statistics, providing coaches and players with actionable performance data previously unavailable from standard video analysis.
\end{abstract}

\newpage
\pagenumbering{arabic} 

\section{Introduction}

In sports analysis systems, precise statistics describing player performance are essential for understanding the game at a larger level. While professional teams can invest significant amounts of money into additional technology and sensors for tracking player performance, other teams may only have a camera for recording games, making the tracking of these statistics more challenging. With only camera data available, a computer vision solution becomes necessary to track players. 

Beyond simply detecting and tracking players, to gain statistical insights such as player speed or distance traveled, the players' coordinate locations must be mapped to their physical locations on the field, so that the known field dimensions can be applied to the players positions to estimate distances. To address this problem, key point detection models can be used to take images as input and then predict where key locations are so that known distances can be linked to the images. Using point homography, we can translate points extracted from the images onto a two-dimensional representation of the field, which can then be used for further post-processing to collect player statistics.

Working with the Milwaukee School of Engineering (MSOE) men’s soccer team, we proposed the creation of a computer vision system that combines these techniques to create a system that analyzes games using only video data. Due to a lack of labeled data, we leveraged pre-trained models such as YOLO and Faster R-CNN to identify players in a given frame. Using these predictions, we prompted Meta's SAM2 with predicted player locations to segment and track players throughout the game footage. To perform key point identification, we trained a custom CNN model on a small dataset of labeled field points to identify these key points on the field and then perform point homography on these identified points. For classifying players into their respective teams, we use clustering techniques as a lightweight method for dividing up players. Combining these components, we can create a video feed of a two-dimensional field with player positioning, separated by team, which can be used for analyzing larger trends about team behaviors and gaining insights from game footage.

With this paper we demonstrate the application of an AI system to create a 2D field representation from video frames in a completely unlabeled and raw dataset. This pattern of keypoint recognition, player detection, team classification, and homography is fairly common in sports analysis systems with AI. However, most of them are applied into a highly curated and labeled dataset. We wanted to replicate this pattern in data that had no kind of preprocessing, labels, or preparation for machine  learning tasks. We show that it is possible to create an automated system that can derive insights from just footage of games. We believe we are paving the way for more flexible and scalable sports analytics solutions, available for the associations with limited resources.


\section{Related Works}

In recent years, advancements in object detection and tracking have significantly improved the ability to analyze player movement in sports applications. Our work leverages previous detection and segmentation models. This section discusses the key methodologies and models that informed our approach.

\textbf{Segment Anything Model 2 (SAM2)}

SAM2 is a transformer architecture with streaming memory for real-time video processing \citep{sam2}. It addresses challenges like occlusions, lighting variations, and temporal consistency by using a memory attention mechanism to store and utilize contextual information across frames. This enables SAM2 to retain object identities and improve segmentation predictions, making it highly effective for long-term tracking applications such as player detection in sports analytics.

\textbf{You Only Look Once (YOLO)}

YOLO (You Only Look Once) revolutionized object detection with its single-stage detection framework, treating the task as a regression problem \citep{yolo}. Unlike traditional models that generate region proposals, YOLO directly predicts bounding boxes and class probabilities from input images in real-time \citep{yoloseries}. YOLOv5 uses CSPDarknet53 as its backbone and incorporates Feature Pyramid Network (FPN) and Path Aggregation Network (PAN) for multi-scale detection \citep{yoloseries}. YOLOv8 builds on YOLOv5 with improvements such as CSPLayer for better accuracy, a new loss function for unbalanced data sets, and improved data augmentation \citep{evolutionyolo}. YOLOv11, the latest version, introduces innovations such as the C3k2 block and C2PSA for enhanced small-object detection and occlusion handling \citep{yolov11}.

These versions were chosen for evaluation due to their advancements in real-time object detection. YOLOv5 is the most commonly used and is widely accessible, making it a reliable choice. YOLOv8 was selected for its direct improvements on YOLOv5 and positive performance in studies like “A Comparative Study of YOLO Series” \citep{yoloseries}. YOLOv11 was included for its use of significantly less parameters while still offering enhanced precision and faster inference due to its state-of-the-art optimizations that are particularly useful for detecting players in complex scenes \citep{yolov11}.

\textbf{Faster R-CNN}

Faster R-CNN is a two-stage object detection model that differs from YOLO's single-stage approach \citep{faster-r-cnn}. While YOLO predicts bounding boxes and class probabilities in one pass, Faster R-CNN first generates region proposals using a Region Proposal Network (RPN), followed by object classification and bounding box refinement \citep{faster-r-cnn}. The architecture includes a backbone for feature extraction, the RPN for generating proposals, and a detection head for classification and refinement \citep{faster-r-cnn}. We chose to evaluate Faster R-CNN to compare its two-stage approach with YOLO’s single-stage detection, providing insights into the strengths and limitations of both frameworks.

Field registration is crucial for converting camera measurements into real-world distances. Traditional methods—such as sensor-based calibration, which fuses data from GPS, accelerometers, and gyroscopes, or template matching that aligns the current view with a stored 2D field representation—often struggle when the field is viewed from non-standard angles or under extreme variations. In soccer, where cameras frequently zoom and shift, these conventional techniques may fail to capture the dynamic nature of the field.

To overcome these challenges, our approach follows a typical sports field registration pipeline: (i) extracting field-specific features (in our case, keypoints), (ii) establishing correspondences between these features and a 2D field template, and (iii) estimating the homography matrix for mapping camera coordinates to real-world distances \citep{Chu_SportsRegistration}. Recent advances in deep learning have shown that CNNs are particularly effective in this domain. For instance, \citep{ZHANG2024685} introduced a cascaded CNN achieving 95.62\% keypoint detection accuracy under challenging conditions, while \citep{Falaleev_keypoint} and \citep{camerafieldregistration} demonstrated the benefits of CNN-based pipelines in enhancing 3D camera calibration and field registration accuracy. Furthermore, optimization-based approaches like PnLCalib and Bayesian frameworks proposed by \citep{gutiérrezpérez2024pnlcalibsportsfieldregistration}, as well as \citep{Claasen_2024}, have significantly improved homography estimation metrics. These developments highlight the power of integrating CNN-based keypoint detection with robust homography estimation to derive precise real-world metrics from raw, dynamic game footage.

\section{Data}

Our primary dataset consists of comprehensive footage from 10 home games during the 2024 season of the MSOE Men’s Soccer team. It captures diverse environmental conditions including daytime, nighttime, and rainy weather scenarios. The video footage was recorded using a high quality elevated BePro camera, providing extensive field coverage while incorporating dynamic movement to track the ball and adjusting zoom levels as needed during gameplay.

\subsection{Player Bounding Boxes}

To generate accurate labels for our dataset, we leveraged the Segment Anything Model (SAM2) for ground truth annotations on 390 consecutive frames. To accelerate the labeling process, we used YOLO predictions to get a inference point to pass as initial prompts to SAM2. While these YOLO predictions had to be manually reviewed, the predictions were often correct, minimizing time manually defining the prompt points for SAM2 when creating a ground truth dataset.


To create bounding box predictions that could then be used for evaluation against objection detection model bounding boxes, we reverse-engineered the SAM2-generated segmentation masks into ground truth bounding boxes using the minimum bounding box rectangle for each mask.\footnote{One challenge presented by this data labeling approach was player entry/departure from the camera frame. This led to a reduction in ground truth points collected for specific players, but the process for dataset collection still allowed for an efficient collection of quality ground truth data}



\subsection{Field Key Points}

Since the original dataset contained no keypoint labeled annotations, the first step in the research involved creating our own ground truth data to support model training and evaluation. We needed to identify and annotate key structural points on the soccer field that could serve as reference points for homography estimation.

\begin{figure}[h]
    \centering
    \includegraphics[width=0.45\textwidth]{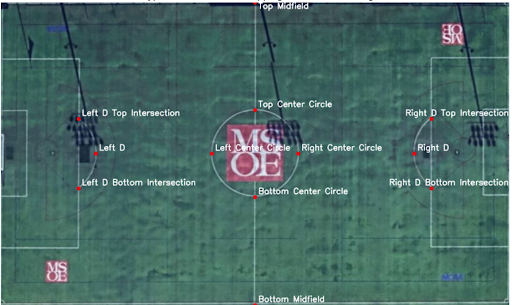} 
    \hspace{0.5cm} 
    \includegraphics[width=0.48\textwidth]{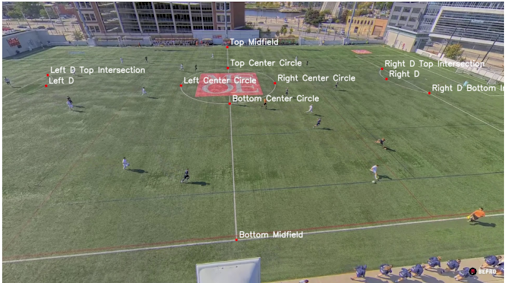} 
    \caption{The set of all keypoints to predict in the field (left) and an example labeled frame from game footage (right).}
    \label{fig:data-keypoints}
\end{figure}

We manually labeled key points across 146 frames extracted from soccer footage, ensuring a diverse representation of field views and camera angles. These frames were extracted from 2 different games, one where there was glare on the field (92 frames), and an overcast day (54 frames), giving us a small dataset that could be evaluated in different weather conditions. Each frame was annotated typically including 4 to 12 per frame depending on visibility  An example of a labeled frame is shown in Figure \ref{fig:data-keypoints} (right).

The selected keypoints were chosen to accurately align each frame with a top-down 2D field template. By focusing on intersections of prominent field markings—such as the penalty arc, center circle, and midfield line—these points serve as strong, consistent visual anchors. Their spatial distribution across the field supports stable homography estimation, while avoiding line extremities helps reduce annotation ambiguity. Additional non-intersecting points were selected to define a perpendicular axis at midfield, further facilitating the transformation. This strategy enables precise alignment for downstream tasks like player localization and tactical analysis.

\section{Methodology}

\begin{figure}[h]
\begin{center}
  \includegraphics[width=0.9\linewidth]{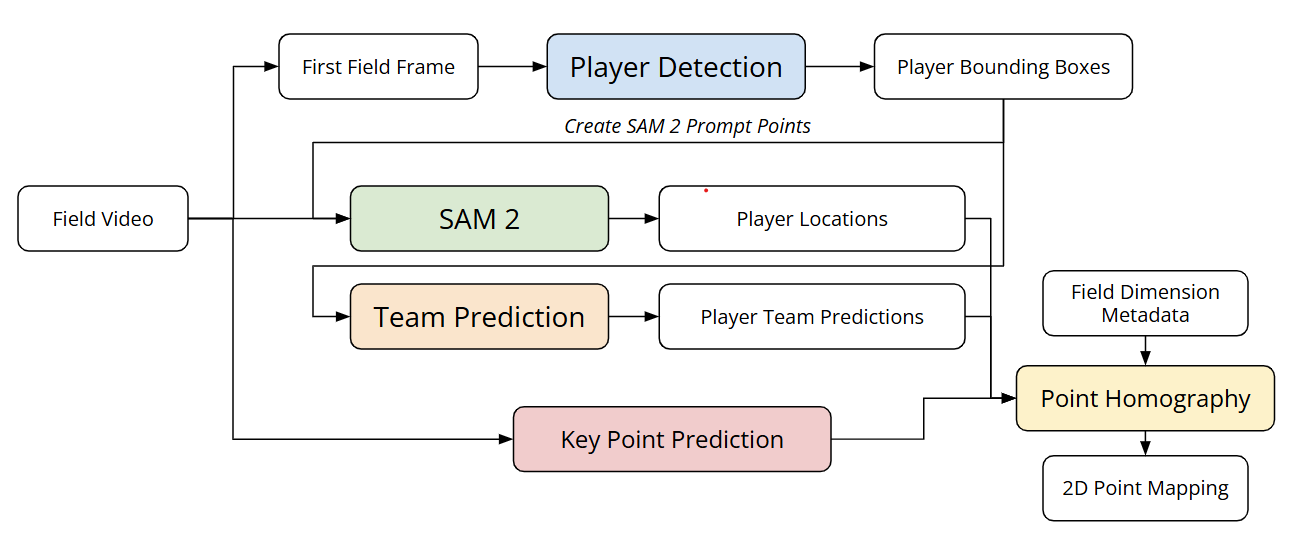}
  \caption{Full workflow for transforming video feed into a 2D representation of the feed. Core model pieces are defined with colors while different states of the data are defined in white.}
  \label{fig:full-system-diagram}
\end{center}
\end{figure}

Since creating a soccer computer vision system is a complex task, we define several workflows to process the game footage to produce meaningful results that can be used for further game analysis, as seen in Figure \ref{fig:full-system-diagram}. The main capabilities we need for a minimal system include: some form of player detection, a way to classify individual players/members of a team, key point predictions to identify the camera's current perspective, and point homography that can transform the frames of game footage onto a 2D representation of the field with physical field measurements. We discuss each of these components further below.

\subsection{Player Detection}

Our approach to player detection and tracking combines an Object Detection model with SAM2 (Segment Anything Model 2). The Object Detector identifies soccer players in the first frame and generates associated bounding boxes. From it we can get the centerpoint of the bounding box for the identified players and use these points as prompts for SAM2, which segments and tracks the players throughout the rest of the video. By leveraging SAM2's unique memory component, we can maintain consistent tracking of individual players, assigning each a unique ID and ensuring continuity even in challenging conditions such as occlusions, players partially moving out of frame, low video resolution, lighting inconsistencies, and varying weather conditions.

 The combination of an Object Detector Model and SAM2 offers several key advantages over traditional approaches like using DeepSORT \citep{deepsort} for tracking. SAM2 provides pixel-accurate tracking and a fine-grained understanding of object shape and position, enabling more precise measurements. Its memory-based design is robust to occlusions and changes in player appearance. Unlike DeepSORT, which relies on continuous detection, SAM2 allows YOLO to be used only once for initialization—after which SAM2 handles the tracking, reducing computational overhead while improving consistency \citep{deepsort}.

Since an Object Detection model alone does not track players across frames, it is used to detect players in the first frame, and its detections serve as input to SAM2 for continuous tracking. These detection models lack ability to associate a detected player in one frame with the same player in the next, which makes them unsuitable for tracking. SAM2 addresses these issues by assigning unique IDs to each segmented player and leveraging memory to track them persistently throughout the video, ensuring robust and reliable player tracking across frames.

The data generated mentioned in section 3.1 will be used as ground truth to measure different Object Detection model performances.

\subsection{Field Key Point Detection}

\begin{figure}[h]
\begin{center}
  \includegraphics[width=0.9\linewidth]{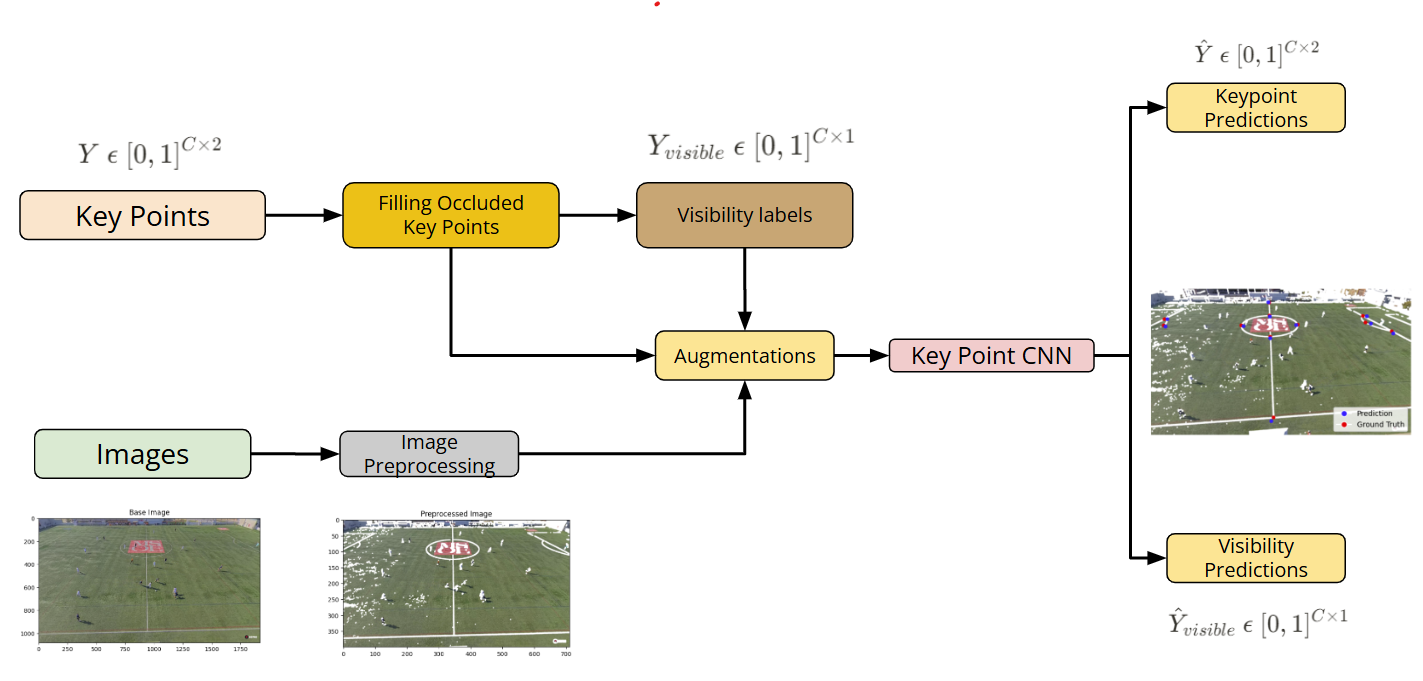}
  \caption{Keypoint prediction model workflow.}
  \label{fig:keyoint-prediction-model}
\end{center}
\end{figure}

\subsubsection{Preprocessing}

To prepare the raw images for training and evaluation of the key point prediction model, we developed a preprocessing pipeline tailored to enhance the visibility of field markings. A thresholding filter was applied to isolate white field lines. The resulting binary mask was then dilated using a morphological kernel to thicken and strengthen the detected field lines. This mask was applied to the original image, and all detected field markings were overwritten with bright white pixels to ensure consistency across samples. The enhanced image was then converted to RGB, resized to 710×400 pixels, to speed up neural network training, and normalized to [0, 1], as seen in the second image of Figure 3.

For key points, original points were processed so that points not in the given frame were imputed with a coordinate of (-1, -1). Additionally, key point coordinates were paired up with a status of their visibility denoted with 0 for points not in the frame and 1 for points in the frame.

\subsubsection{Augmentation}

To improve model robustness and dataset diversity, we applied horizontal flipping for data augmentation. Each frame was flipped and its keypoint annotations adjusted by inverting the normalized x-coordinates $(1 - x)$ and swapping left/right labels. This encouraged the model to focus on field line patterns rather than extraneous features, effectively doubling the dataset while preserving spatial structure. The dataset was split 80/20 for training and testing.

\subsubsection{Model Architecture}

We designed a multi-task convolutional neural network to jointly predict the visibility and location of 12 predefined soccer field keypoints. The model outputs a visibility probability (between 0 and 1) and normalized coordinates (also in the range [0, 1]) for each keypoint.


 To manage keypoints that are not visible due to occlusion or camera angles, we employ a custom loss function that calculates the mean absolute error (MAE) only over visible keypoints (ignoring those labeled as (-1, -1)) and binary cross-entropy loss for visibility (defined in Equation \ref{eq:masked-mae}). These losses are combined with weights of 10 for the coordinate loss and 1 for the visibility loss.

 We trained this network using the Adam Optimizer with an exponential decay learning rate schedule. Additionally, to find a strong set of model hyperparameters, we perform a Bayesian Optimization hyperparameter search over factors including dropout rate, layer size, and learning rate. An example output of the model can be seen in the right side of Figure \ref{fig:keyoint-prediction-model}.


\subsection{Homography}

In terms of the homography, the real life dimensions of the field were collected with the use of Google Maps API. Combined with official NCAA collegiate soccer rules for measurements on the field, we defined all keypoint locations on the 2D field mapping. This image and dimensions were used to align predicted key points from each video frame with a standardized 2D representation of a soccer field; we estimated a homography transformation based on a set of corresponding points. 

For the calculation of the homography matrix we used different algorithms. The most successful one was the Direct Linear Transformation (DLT), commonly used in the homography field \citep{homography_KANG201468}. The DLT estimates the homography matrix by solving a linear system built from point correspondences, using Singular Value Decomposition. It assumes all input points are accurate and does not account for outliers, making it highly sensitive to keypoint placement.

Once we obtained the homography matrix between the 3D and 2D view, any point in the image could be converted from one to another by applying the matrix to it. This gave us what we needed to jump from the camera perspective to a top down angle, in which real world metrics and measures can be taken. 

\subsection{Player Mapping and Team Recognition}

\begin{figure}[h]
\begin{center}
  \includegraphics[width=0.7\linewidth]{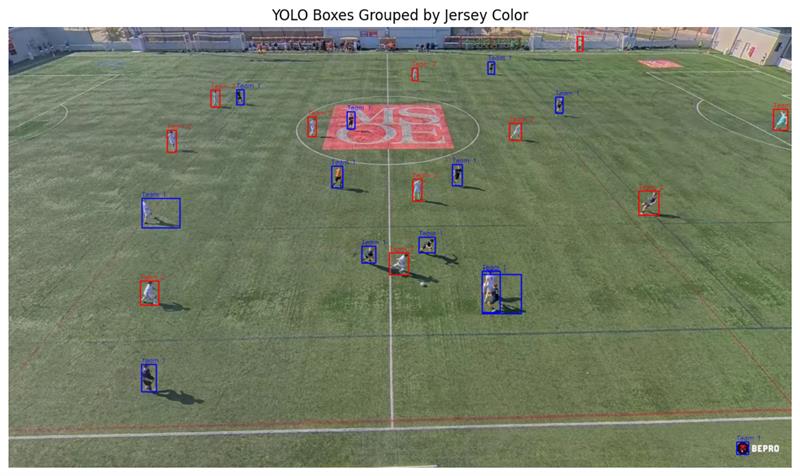}
  \caption{Team assignment based on pixel color clustering of player bounding boxes. This example uses player detections from YOLOv8.}
  \label{fig:team-prediction}
\end{center}
\end{figure}

Once we had both parts separate, player detection and homography, we tried to connect them and utilize clustering to classify each of the players into a team. For each person detected by the model, we computed the average color within a small patch ($5\times5$ pixels) centered at the middle of the bounding box, ensuring robustness to noise and avoiding background influence. The resulting color vectors, represented in RGB space, were then clustered using the K-Means algorithm. We used $k=2$ to detect both teams involved in a soccer game. Each bounding box was then assigned to a cluster based on its color similarity, allowing us to group players by team without requiring manual labeling. This process served as a lightweight and effective method for unsupervised team classification based on visual appearance, as demonstrated in Figure \ref{fig:team-prediction}. 

\section{Results}

\subsection{Player Detection Model Performance}

In evaluating different object detection models for identifying soccer players as prompts for SAM2, F1-score, IoU (Intersection over Union), recall, and precision each play a crucial role in determining effectiveness. Of these metrics, F1-score is the most valuable in our application, as it balances both precision and recall. Recall was deemed the next most critical metric since missed detections means that players will not be passed to SAM2 for segmentation, limiting tracking completeness. Precision, while important, is slightly less critical in our application because false positives outside the typical player size are unlikely to result in valid segmentation by SAM2 since we are filtering by the size of the mask. Finally, IoU is used to measure the alignment of the bounding box, but small misalignments are less problematic if the object is still successfully segmented. Based on these priorities and the comparative results shown in this results section, we have chosen to pursue YOLOv5x as the best model for our application. As seen in Table 1,  it offers the highest F1-score (0.8451) while maintaining strong recall (0.7995), precision (0.8963), and IoU (0.7644), making it highly reliable for consistent player detection and optimal integration with SAM2. 

We evaluated four models: Faster R-CNN, YOLOv5x, YOLOv8x, and YOLOv11x. Each model was tested on its ability to generate accurate bounding boxes that identify players on the soccer field  to serve as prompts to SAM2. Table 1 represents the results of these key performance metrics.  Note that in each version, the "extra-large" model was tested to ensure a fair assessment. 

\begin{table}[h]
    \centering
    \begin{tabular}{|l|c|c|c|c|}
        \hline
        \textbf{Model} & 
        \textbf{Mean IoU} & 
        \textbf{Recall} & 
        \textbf{Precision} & 
        \textbf{F1-Score} \\ 
        \hline
        Faster R-CNN  & 0.6837  & 0.6574 & 0.7942 & 0.7194  \\ 
        YOLOv11x  & 0.7934  & 0.6805 & 0.9229 & 0.7834    \\ 
        YOLOv8x  & 0.7466  & 0.8152  & 0.8765 & 0.8447   \\ 
        YOLOv5x& 0.7644  & 0.7995  & 0.8963 & 0.8451   \\ 
        \hline
    \end{tabular}
    \caption{Comparison of different object detection models on SAM2 generated ground truth samples.}
    \label{tab:object-detection-eval}
\end{table}

YOLOv5 indicated the best balance between precision and recall. YOLOv8x followed closely behind with a slightly higher recall but slightly lower precision. YOLOv11x had the highest IoU and precision, but its recall was significantly lower, which can lead to missed detections, which is crucial in our application. All three YOLO models tested outperformed Faster R-CNN across every key metric, demonstrating the superior effectiveness of modern YOLO architectures in this application.

Using YOLOv5x, we successfully identified 17/22 players without any fine-tuning. These players initially identified were successfully segmented and tracked through SAM2, where their positions and movements are accurately monitored across sequential frames. Despite minor exceptions with player re-identification and identifying fans, the pipeline demonstrates promising performance. With potential YOLO fine-tuning, the tracking process can be further optimized, leading to a more robust and reliable player tracking system.

\subsection{Keypoint Prediction Model Performance}

We defined tailored metrics for the evaluation of our CNN. For the evaluation of the key points prediction we used a masked Mean Absolute Error (MAE) as described in Equation \ref{eq:masked-mae}, which computed the error only for visible keypoints, ensuring key points that are not in the frame (denoted by a special value) do not negatively impact the model’s performance evaluation. This approach provides a more reliable assessment of localization accuracy by focusing on keypoints with valid ground truth data.



\begin{align}
MAE_{masked} = \frac{1}{N}\sum^{N}_{i=1}\frac{\sum^{C}_{j=1}v_{i,j}(|x_{i,j}-\hat{x}_{i,j}| + |y_{i,j} - \hat{y}_{i,j}|)}{\sum^{C}_{j=1}v_{i,j}}\label{eq:masked-mae}
\end{align}

Meanwhile, accuracy was employed to evaluate visibility predictions by comparing the rounded model outputs to the ground truth labels as described in Equation \ref{eq:rounded-accuracy}. This metric is particularly useful in binary classification tasks, such as determining whether a keypoint is visible or not. Together, these metrics provide a comprehensive evaluation of both the spatial accuracy of the detected keypoints and the model’s ability to correctly classify their visibility, ensuring robust and interpretable performance analysis.



\begin{align}
Accuracy = \frac{1}{N}\sum^{N}_{i=1}\frac{1}{C}\sum^{C}_{j=1}[round(\hat{y}_{i,j}) = y_{i,j}]\label{eq:rounded-accuracy}
\end{align}

As shown in table 2, the proposed convolutional neural network demonstrated strong performance across both keypoint visibility classification and coordinate regression tasks. On the training set, the model achieved a visibility accuracy of 99.89\%, with still high accuracy of 97.18\% on the validation set, indicating strong generalization and minimal overfitting. For keypoint localization, the model achieved a mean absolute error (MAE) of 0.0107 in normalized image coordinates on the training set and 0.0138 on the validation set, corresponding to pixel errors of 5.96 and 7.65 pixels, respectively. These results suggest that the model is capable of precisely identifying and localizing field keypoints under varying visual conditions, providing a reliable foundation for downstream tasks such as homography estimation.

\begin{table}[h]
    \centering
    \begin{tabular}{|l|c|c|}
        \hline
        \textbf{Metric} & 
        \textbf{Training Set} & 
        \textbf{Testing Set} \\ 
        \hline
        Accuracy  & 0.9989  & 0.9718  \\ 
        MAE (\% image)  & 0.0107  & 0.0138    \\ 
        MAE (pixels)  & 5.9648  & 7.6531   \\ 
        \hline
    \end{tabular}
    \caption{Results of the keypoint prediction model on training and testing sets.}
    \label{tab:keypoint-detection-eval}
\end{table}

\subsection{Full System Analysis}

\begin{figure}[h]
\begin{center}
  \centering
    \includegraphics[width=0.45\textwidth]{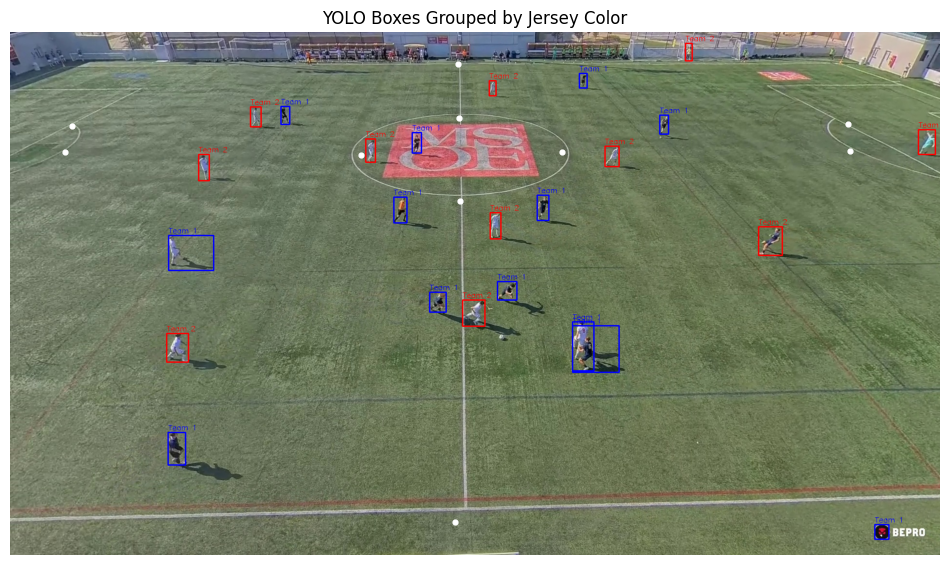} 
    \hspace{0.5cm} 
    \includegraphics[width=0.5\textwidth]{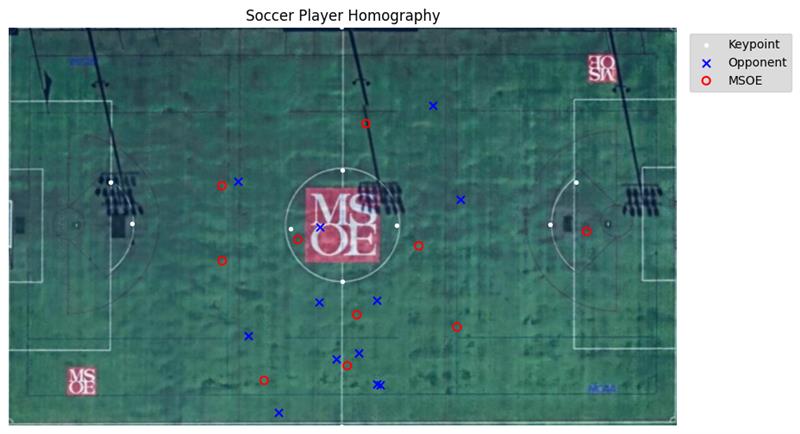} 
  \caption{An example frame with player detection/keypoint predictions (left) and then applying homography to create a 2D field representation (right).}
  \label{fig:team-prediction-homography}
\end{center}
\end{figure}

In order to measure the performance of the system as a whole, the integration of the homography and keypoint detection, we calculated the MAE and projection errors with the actual data. Our MAE of the actual keypoints was 0.225 meters for the ground truth points and 0.26 meters for the predicted points. When comparing the homography of the predicted vs the ground truth, we obtained an average projection error of 0.499 meters in the keypoints transformed.

Looking at the results qualitatively, from a general perspective, there are several areas of improvement. As we can see in figure 5, our player detection model suffers from false positives, such as a ballboy being recognized or a referee. As well, the clustering algorithm mistakes players when they are affected by the glare or shadows, assigning them to the opposite team. Furthermore, our keypoint detection model is not perfect, which can lead to innaccurate positions in the top down perspective or misleading stats. These are some of the challenges that we are facing at the moment, and that make the system not completely  ready yet.

\section{Future Work}


In future work, we plan to improve player tracking by addressing the problem of re-identification for players entering and leaving the frame. Using YOLO, we can maintain player counts and identify and track these players as they exit and re-enter the frame. When a player returns, we aim to re-prompt SAM2 using their previous YOLO coordinates, focusing only on re-prompting that player. By maintaining specific player attributes like the player’s team and their exit location when leaving the frame, we can ensure consistent player identification, even after they leave and re-enter the field. Additionally, being able to track single players across the entirety of the game would allow for stat estimation at the player-level, rather than at the team-level as our system currently can do.

Additionally, for our trained parts of the system (specifically our key point detection model), we have only collected our dataset from game footage from MSOE’s home field, potentially making our models overfit to the camera and field at MSOE. This makes the system less advantageous when there are away games that we want to analyze statistics from. To improve on this, we would want to augment our dataset with samples from other fields with different camera angles and backgrounds so the trained portions of our models generalize across all fields. As well, we will work on ball detection in order to create statistics related to the actual game flow.

\section{Conclusion}

In this work, we introduced our computer vision system for extracting a 2D field representation of players from soccer game footage. Because of a lack of data for more complex tasks like player detection, we leverage pretrained foundation models such as YOLO and SAM2 to classify players and segment the video based on the predicted player locations, respectively. We use a clustering approach based on predicted player bounding boxes to classify teams without needing to develop a supervised classifier. To connect predicted players and team affiliations to their corresponding physical distances, we train a custom CNN model on a limited dataset of key points, so we may predict these points and perform homography of known locations in frames to a consistent two-dimensional plane. Combining all these pieces, we can evaluate team statistics over the course of a game using only camera footage. While there are more statistics that can be computed based on these initial outputs our computer vision system generates, we hope that these initial results can begin to be used by the Milwaukee School of Engineering soccer teams to gain further insights about player activity during games.

\section{Acknowledgments}

The authors would like to thank coach Rob Harrington of the MSOE men’s soccer team for providing the team with access to game footage from the Milwaukee School of Engineering men’s soccer games and practices, which were used to create the data sets described above. The authors would also like to acknowledge Nicolas Picha, Joseph Loduca, Asher Harris, and Patrick Marshall from the MSOE men’s soccer team for their collaboration with the authors in discussing this work.


\bibliography{resources}

\end{document}